# ChrEnTranslate : Cherokee-English Machine Translation Demo with Quality Estimation and Corrective Feedback

**Shiyue Zhang    Benjamin Frey    Mohit Bansal**
UNC Chapel Hill
{shiyue, mbansal}@cs.unc.edu; benfrey@email.unc.edu

## Abstract

We introduce ChrEnTranslate, an online machine translation demonstration system for translation between English and an endangered language Cherokee. It supports both statistical and neural translation models as well as provides quality estimation to inform users of reliability, two user feedback interfaces for experts and common users respectively, example inputs to collect human translations for monolingual data, word alignment visualization, and relevant terms from the Cherokee-English dictionary. The quantitative evaluation demonstrates that our backbone translation models achieve state-of-the-art translation performance and our quality estimation well correlates with both BLEU and human judgment. By analyzing 216 pieces of expert feedback, we find that NMT is preferable because it copies less than SMT, and, in general, current models *can translate fragments of the source sentence but make major mistakes*. When we add these 216 expert-corrected parallel texts back into the training set and retrain models, equal or slightly better performance is observed, which indicates the potential of human-in-the-loop learning.[1]

## 1 Introduction

Machine translation is a relatively mature natural language processing technique that has been deployed to real-world applications. For instance, Google Translate currently supports translations of over 100 languages. However, a lot of low-resource languages are out there without the support of modern technologies, which might accelerate their vanishing. In this work, we focus on one of those languages, Cherokee. Cherokee is one of the most well-known Native American languages, however, is identified as an "endangered" language by UNESCO. Cherokee nations have carried out language revitalization plans (Nation, 2001) and established language immersion programs and k-12 language curricula. Cherokee language courses are offered in some universities, including UNC Chapel Hill, the University of Oklahoma, Stanford University, Western Carolina University. A few pedagogical books have been published (Holmes and Smith, 1976; Joyner, 2014; Feeling, 2018) and a digital archive of historical Cherokee language documents has been built up (Bourns, 2019; Cushman, 2019). However, there are still very limited resources available on the Internet for Cherokee learners; meanwhile, first language speakers and translators of Cherokee are mostly elders and would likely benefit from machine translation's assistance. This motivates us to develop the first online Cherokee-English machine translation demonstration system. Extending our previous works (Frey, 2020; Zhang et al., 2020), we develop the backbone statistical and neural machine translation systems (SMT and NMT) on a larger parallel dataset (17K) and obtain the state-of-the-art Cherokee-English (Chr-En) and English-Cherokee (En-Chr) translation performance.

Besides translation, our system also supports quality estimation (QE) for both SMT and NMT. QE is an important (missing) component of machine translation systems, which is used to inform users of the reliability of machine-translated content (Specia et al., 2010). Since our models are trained on a very limited number of parallel sentences, it is expected that the translations will be poor in most cases when used by Internet users. Therefore, QE is essential for avoiding misuse and warning users of potential risks. Existing best-performance QE models are usually trained under supervision with quality ratings from professional translators (Fomicheva et al., 2020a). However, we are unable to easily collect a lot of human ratings for Cherokee, due to its state of endangerment. Nonetheless, we test both supervised and unsupervised QE methods: (1) *Supervised*: we use BLEU (Papineni et al., 2002) as the quality rating proxy and train a BLEU regressor; (2) *Unsupervised*: following the uncertain estimation literature (Lakshminarayanan et al., 2017), we use the ensemble model's output probability as the estimation of quality. Furthermore, to evaluate how well the QE models perform, we collect 200 human quality ratings (50 ratings for SMT Chr-En, SMT En-Chr, NMT Chr-En, and NMT En-Chr, respectively). We show that our methods obtain moderate to strong correlations with human judgment (Pearson correlation coefficient $\gamma \geq 0.44$).

---
[1] Our online demo is at https://chren.cs.unc.edu/; our code is open-sourced at https://github.com/ZhangShiyue/ChrEnTranslate; and our data is available at https://github.com/ZhangShiyue/ChrEn.

One main purpose of our system is to allow human-in-the-loop learning. Since limited parallel texts are available, it is important to involve humans, especially experts, in the loop to give feedback and then improve the models accordingly. We develop two different user feedback interfaces for experts and common users, respectively (shown in Figure 2). We ask experts to provide quality ratings, correct the model-translated content, and leave open-ended comments; for common users, we allow them to rate how helpful the translation is and to provide open-ended comments. Upon submission, we collected 216 pieces of feedback from 4 experts. We find that experts favor NMT more than SMT because SMT excessively copies from source sentences; according to their ratings and comments, current translation systems *can translate fragments of the source sentence but make major mistakes*. Our naive human-in-the-loop learning, by adding these 216 expert-corrected parallel texts back to the training set, obtains equal or slightly better translation results. Plus, the expert comments shine a light on where the model often makes mistakes. Besides, our demo allows users to input text or choose an example input to translate (shown in Figure 1). These examples are from our monolingual databases so that experts will annotate them by providing translation corrections. Finally, to support an intermediate interpretation of the model translations, we visualize the word alignment learned by the translation model and link to cherokeedictionary to provide relevant terms from the dictionary.

Our code is hosted at ChrEnTranslate and our online website is at chren.cs.unc.edu. Common users need to accept agreement terms before using our service to avoid misuse; access the expert page chren.cs.unc.edu/expert requires authorization. We encourage fluent Cherokee speakers to contact us and contribute to our human-in-the-loop learning procedure. A demonstration video of our website is at YouTube. In summary, our demo is featured by (1) offering the first online machine translation system for translation between Cherokee and English, which can assist both professional translators or Cherokee learners; (2) documenting human feedback, which, in the long run, expands Cherokee data corpus and allows human-in-the-loop model development. Additionally, our website can be easily adapted to any other low-resource translation pairs.

## 2 System Description

### 2.1 Translation Models

As shown in Figure 1, our system allows users to choose the statistical or neural model (SMT or NMT).

**SMT** is more effective for out-of-domain translation between Cherokee and English (Zhang et al., 2020). We implement phrase-based SMT model via Moses (Koehn et al., 2007), where we train a 3-gram KenLM (Heafield et al., 2013) and learn word alignment by GIZA++ (Och and Ney, 2003). Model weights are tuned on a development set by MERT (Och, 2003).

**NMT** has better in-domain performance and can generate more fluent texts. We implement the global attentional model proposed by Luong et al. (2015). Detailed hyper-parameters can be found in Section 3.1. Note that we do not use Transformer because it empirically works worse (Zhang et al., 2020). And we find that the multilingual techniques we explored only significantly improve in-domain performance when using multilingual Bible texts, so we suspect that it biases to Bible-style texts. Hence, we also do not apply multilingual techniques and just train the backbone models with our Cherokee-English parallel texts. We use a 3-model ensemble as our final working model.

### 2.2 Quality Estimation

**Supervised QE.** The QE (Specia et al., 2010) task in WMT campaign provides thousands of model-translated texts plus corresponding human ratings, which allow participants to train supervised QE models. Fomicheva et al. (2020a) show that supervised models work significantly better than unsupervised ones. Since we are unable to collect thousands of human ratings, we use BLEU (Papineni et al., 2002) as the quality rating. We use 17-fold cross-validation to obtain training data, i.e., we split our 17K parallel texts into 17 folds, use 16 folds to train a translation model, get the translation features plus BLEU scores of examples in the left one fold, repeat this for 17 times, and finally, we get the features plus BLEU scores of 17K examples. Then, we separate 16K examples as a training set to train a BLEU score regressor and evaluate the performance on the left 1K examples. Fomicheva et al. (2020a,b) define three sets of features. However, we need to compute features online, so some features (e.g., dropout features) that require multiple forward computations will greatly increase latency. W use features that will not cause too much speed lag. For SMT, we use:

(a) output length $L_t$, i.e., the number of words in the translated text;

(b) total score;

(c) scores of distortion, language model, lexical reordering, phrases penalty, translation model, and word penalty;

(d) length normalized (b) and (c) features (i.e., divide each feature from (b) and (c) by (a)).

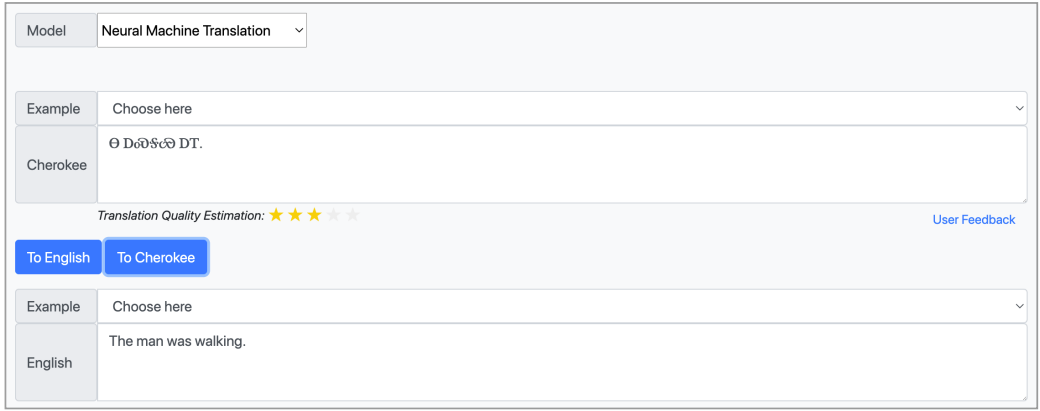

Figure 1: Translation interface of our demonstration system. Note that "Ꭴ ᎠᏯᏍᎤ DT." is not a correct translation. See Figure 2 for the corrected translation by an expert.

For NMT, we use:

(a) output length;

(b) log probability and length normalized log probability;

(c) probability and length normalized probability;

(d) attention entropy (Fomicheva et al., 2020a,b): $-\frac{1}{L_t} \sum_{i=1}^{L_t} \sum_{j=1}^{L_s} \alpha_{ij} \log \alpha_{ij}$, where $L_s$ is the length of source text, and $\alpha_{ij}$ is the attention weight between target token $i$ and source token $j$.

Finally, we use XGBoost (Chen and Guestrin, 2016) as the BLEU regressor.[2] As shown in Figure 1, we use 5 stars to show QE, therefore, we rescale the estimated quality to 0-5 by dividing the predicted BLEU score (0-100) by 20.

**Unsupervised QE.** Even though supervised QE works better (Fomicheva et al., 2020a), we suspect that the advantage cannot generalize to open domain scenarios unless we have a large amount of human-rated data to learn from. Hence, we also explore unsupervised QE methods. Unsupervised QE is closely related to uncertainty estimation. We can use how uncertain the model is to quantify how low-quality the model output is. Though it is intuitive to use the output probability as the model's confidence, Guo et al. (2017) point out that the output probability is often poorly calibrated, so that they propose to re-calibrate the probability on the development set. However, this method is designed for classification tasks and is not applicable for language generation. Gal and Ghahramani (2016) show that "dropout" can be a good uncertainty estimator, inspired by which Fomicheva et al. (2020b) propose the dropout features. However, the multiple forward passes are not preferable for an online system. Lakshminarayanan et al. (2017) demonstrate that the ensemble model's output probability can better estimate the model's uncertainty than dropout. We find that this method is simple yet effective for NMT. Note that we normalize the output probability by the sentence length. Similarly, we rescale the normalized probability (0-1) to 0-5 by multiplying it by 5.

---

[2]We also tested GradientBoost (Friedman, 2002) and MLP, but XGBoost empirically works better.

Figure 2: Two user feedback interfaces of our demonstration system. (b) shows the feedback given by an expert.

**Human Quality Rating.** So far, our QE development and evaluation are all based on BLEU. To better evaluate QE performance, we collect 200 human ratings (all rated by Prof. Benjamin Frey[3]), 50 ratings for Chr-En SMT, En-Chr SMT, Chr-En NMT, and En-Chr NMT, respectively. We follow the direct assessment setup used by FLoRes (Guzmán et al., 2019),[4] and thus each translated sentence receives a 0-100 quality rating.

---
[3]Benjamin Frey is a proficient second-language Cherokee speaker and a citizen of the Eastern Band of Cherokee Indians.

[4]0–10: represents a translation that is completely incorrect and inaccurate; 11–29 represents a translation with a few correct keywords, but the overall meaning is different from the source; 30–50 represents a translation that contains translated fragments of the source string, with major mistakes; 51–69 represents a translation that is understandable and conveys the overall meaning of source string but contains typos or grammatical errors; 70–90 represents a translation that closely preserves the semantics of the source sentence; 90–100 range represents a perfect translation.

### 2.3 User Feedback & Example Inputs

Enlarging the parallel texts is a fundamental approach to improve the translation model's performance. Besides compiling existing translated texts, it is important to newly translate English texts to Cherokee by translators. Our system is designed to not only assist these translators but also document their feedback and post-edited correct translation, so that model can be improved by using this feedback, i.e., human-in-the-loop learning. To achieve this goal, we design two kinds of user feedback interfaces. One is for common users, in which users can rate how helpful the translation is (in 5-point Likert scale) and leave open-ended comments, as shown in Figure 2 (a). The other is for experts, in which authorized users can rate the quality, correct the translated text, and leave open-ended comments, as shown in Figure 2 (b). Upon submission, we collect 216 pieces of feedback from 4 experts and detailed analysis can be found in Section 3.3. Meanwhile, as shown in

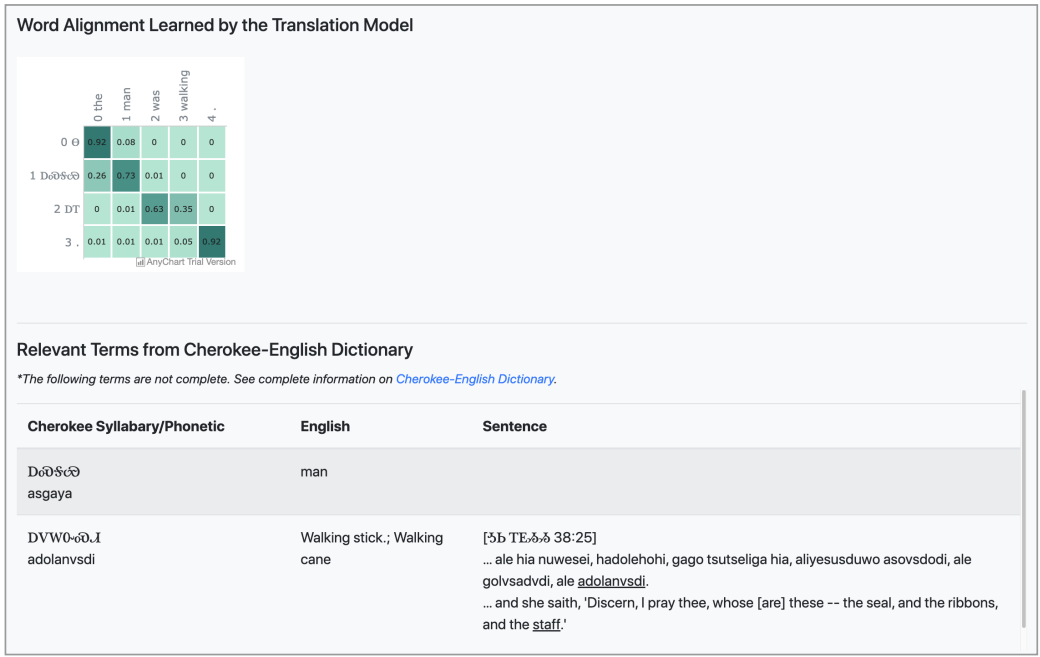

Figure 3: Word alignment visualization and link to Cherokee-English Dictionary.

Figure 1, besides inputting text, users can also choose an example input to translate. These examples are from our Cherokee or English monolingual databases. On the one hand, this provides users with more convenience; on the other hand, whenever experts submit translation corrections of an example, we will update its status as "labeled". Hence, we can gradually collect human translations for the monolingual data.

### 2.4 Other Features

As shown in Figure 3, to make model prediction more interpretable to users, we **visualize the word alignment** learned by the translation model. For SMT, we visualize the hard word-to-word alignment; for NMT, we visualize the soft attention map between source and target tokens. Additionally, to provide users with some oracle and handy references from the dictionary, we **link to cherokee-dictionary**. We use each of the source and target tokens as a query and list up to 15 relevant terms on our web page.

## 3 Evaluation

### 3.1 Implementation Details

**Data.** To train translation models, we use the 14K parallel data collected by our previous work (Zhang et al., 2020) plus 3K newly complied parallel texts. We randomly sample 1K as our development set and treat the rest as the training set. The data is open-sourced at ChrEn/data/demo. To collect human quality ratings, we randomly sample 50 examples from the development set, and for each of them, we collect 4 ratings for Chr-En/En-Chr SMT and Chr-En/En-Chr NMT, respectively.

**Setup.** We implement SMT models via Moses (Koehn et al., 2007). After training and tuning, we run it as a server process.[5] We develop our NMT models via OpenNMT (Klein et al., 2017). For both Chr-En and En-Chr NMT models, we use 2-layer LSTM encoder and decoder, general attention (Luong et al., 2015), hidden size=1024, label smoothing (Szegedy et al., 2016) equals to 0.2, dynamic batching with 1000 tokens. Differently, the Chr-En NMT model uses dropout=0.3, BPE tokenizer (Sennrich et al., 2016), and minimum word frequency=10; the En-Chr NMT model uses dropout=0.5, Moses tokenizer, and minimum word frequency=0. We train each NMT model with three random seeds (7, 77, 777) and use the 3-model ensemble as the final translation model, and we use beam search (beam size=5) to generate translations. We implement the supervised QE model with XGBoost.[6] XGBoost has three important hyperparameters: max depth, eta, the number of rounds. Tuned on the development set, we set them as (5, 0.1, 100) for Chr-En SMT, (3, 0.1, 80) for En-Chr SMT, (4, 0.5, 40) for Chr-En NMT, and (5, 0.1, 40) for En-Chr NMT. Lastly, the backend of our demonstration website is based on the Flask framework.

---

[5] http://www.statmt.org/moses/?n=Advanced.Moses
[6] https://xgboost.readthedocs.io/en/latest/python/index.html

|       |              |                          | BLEU  |       | Human Rating |       |
|-------|--------------|--------------------------|-------|-------|--------------|-------|
| Model |              | QE                       | Chr-En | En-Chr | Chr-En | En-Chr |
| SMT   | Supervised   | XGBoost                  | **0.75** | **0.71** | **0.63** | **0.44** |
|       | Unsupervised | TranslationModel / length | 0.36  | 0.46  | 0.07  | -0.09 |
|       |              | LM / length              | 0.34  | 0.43  | -0.11 | 0.11  |
|       |              | PhrasePenalty / length   | -0.33 | -0.52 | 0.06  | 0.03  |
| NMT (ensemble) | Supervised | XGBoost            | **0.79** | **0.68** | 0.53 | 0.38 |
|       | Unsupervised | Exp(LogProbability / length) | 0.75 | 0.63 | **0.59** | 0.44 |
|       |              | LogProbability / length  | 0.45  | 0.50  | 0.37  | **0.52** |

Table 1: Pearson correlation coefficients between QE and BLEU or between QE and human rating. "/ length" represents the normalization by output sentence length.

| Model | Chr-En | En-Chr |
|-------|--------|--------|
| SMT   | 17.0   | 12.9   |
| NMT (single) | 18.1 | 13.8 |
| NMT (ensemble) | **19.9** | **14.8** |

Table 2: The performance of translation models.

**Metrics.** We evaluate translation systems by BLEU (Papineni et al., 2002) calculated via Sacre-BLEU[7] (Post, 2018). Supervised QE models are developed by minimizing the mean square error of predicting BLEU, but all QE models are evaluated by the correlation with BLEU on development set and the correlation with human ratings. We use Pearson correlation (Benesty et al., 2009).

### 3.2 Quantitative Results

**Translation.** Table 2 shows the translation performance on our 1K development set, which is significantly better than the single-model in-domain translation performance reported in our previous work (Zhang et al., 2020) and thus achieves the state-of-the-art results. In addition, the 3-model NMT ensemble further boosts the performance.

**QE.** Table 1 illustrates the performance of quality estimation models. In our experiments, we take every feature used in supervised QE as an unsupervised quality estimator. Here, we only present those having a high correlation with BLEU and human rating. It can be observed that, for SMT, supervised QE consistently works better, whereas, for NMT, unsupervised QE has a better correlation with human rating. The obtained correlations with human judgement are moderate ($\gamma \geq 0.3$) to strong ($\gamma \geq 0.5$) (Cohen, 1988). Therefore,

---

[7]BLEU+c.mixed+#.1+s.exp+tok.13a+v.1.5.0

---

we use the trained XGBoost for SMT model's QE and use the length normalized probability (i.e., Exp(LogProbability / length)) for NMT model's QE in our online demonstration system.

### 3.3 Qualitative Results

**Expert Feedback.** Upon submission, we received 216 pieces of feedback from 4 experts (including Prof. Benjamin Frey and 3 other fluent Cherokee speakers). The results are shown in Table 3. It can be observed that we received a lot more feedback to NMT than SMT because SMT excessively copies words from source sentences when translating open-domain texts whereas NMT can mostly translate into the target language. On average, there are only 2.3 tokens in the input or translated Cherokee sentence; however, the average translation quality rating is only 2.45 out of 5, which is close to the average rating (43.8 out of 100) of the 200 human ratings we collected. Therefore, according to FLoRes's rating standard (Guzmán et al., 2019) (see footnote 2), our translation systems *can translate fragments of the source string but make major mistakes* in general. Besides ratings, we received 36 open-ended comments that shine a light on common mistakes made by the models. The most frequent comments are (1) *model gets some parts correct but others wrong*. For example, "it got the subject but not the verb", "it got the stem right but used 3rd person prefix", "it missed the part about going to town, but got 'today' correct", etc. (2) *model uses archaic English terms*, like "thy", "thou", "speaketh", etc. because the majority of our training set is the Cherokee Old Testament and the Cherokee New Testament.

**Human-in-the-Loop Learning.** To improve models based on expert feedback, we propose to simply add the 216 expert-corrected parallel texts back to our training set and retrain the translation models.[8] The new BLEU results on our development set are 17.3, 13.0, 20.0, 14.8 for Chr-En SMT, En-Chr SMT, Chr-En NMT (ensemble), and En-Chr NMT (ensemble), respectively, which are equal or slightly better than the results in Table 2. To tackle the archaic English issue, we simply replace archaic English terms ("thy", "thou") with new English terms ("your", "you").

| Model | Chr-En | En-Chr |
|---|---|---|
| SMT | 12 / 1.92 / 0.39 | 6 / 2.0 / 0.66 |
| NMT | 166 / 2.58 / 0.43 | 32 / 2.13 / 0.21 |

Table 3: Expert feedback. In each cell, the 3 numbers are the number of feedback received / average quality rating / Pearson correlation coefficient between quality rating and quality estimation.

## 4 Conclusion & Future Work

In this work, we develop a Cherokee-English Machine Translation demonstration system that intends to demonstrate and support automatic translation between Cherokee and English, collect user feedback/translations, allow human-in-the-loop development, and eventually contribute to the revitalization of the endangered Cherokee language. Future work involves inviting more experts and common users to test/use our system and proposing more efficient and effective human-in-the-loop learning methods.

## 5 Broader Impact Statement

As shown in Section 3.3, the current translation models are still far from being reliably used in practice. Therefore, our system is just a demonstration or prototype of the translation between Cherokee and English, while the model-translated texts are not supposed to be directly applied anywhere else without confirmation from professional translators. We stress this point in our agreement terms. Common users need to accept those terms before using our system; experts need to agree to those terms as well before being authorized. Lastly, we sincerely thank David Montgomery, Barnes Powell, and Tom Belt for voluntarily participating in our system test and providing their feedback.

## Acknowledgments

We thank the reviewers for their helpful comments, and we thank David Montgomery, Barnes Powell, and Tom Belt for voluntarily participating in our system test and providing their valuable feedback. This work was supported by NSF-CAREER Award 1846185, ONR Grant N00014-18-1-2871, and a Microsoft Investigator Fellowship. The views contained in this article are those of the authors and not of the funding agency.

---

[8]We also tried to up-weight these examples by repeating them by 5 or 10 times but did not see better performance.